\pdfoutput=1

\documentclass[11pt]{article}

\usepackage{EMNLP2023}

\usepackage{times}
\usepackage{latexsym}
\usepackage[utf8]{inputenc}
\usepackage{graphicx}
\usepackage{array}
\usepackage{threeparttable}
\usepackage{fmtcount}

\usepackage[T1]{fontenc}

\usepackage[utf8]{inputenc}

\usepackage{microtype}

\usepackage{inconsolata}

\title{
Post Turing: \\ Mapping the landscape of LLM Evaluation}

\author{   Alexey Tikhonov\\
Inworld.AI \\
Berlin, Germany \\
\texttt{altsoph@gmail.com} \And
Ivan P. Yamshchikov \\
  CAIRO, THWS\\
 W\"urzburg, Germany\\
 CEMAPRE, ISEG, \\
 University of Lisbon, Portugal\\
  \texttt{ivan@yamshchikov.info} \\}

\begin{document}

\maketitle
\begin{abstract} In the rapidly evolving landscape of Large Language Models (LLMs), introduction of well-defined and standardized evaluation methodologies remains a crucial challenge. This paper traces the historical trajectory of LLM evaluations, from the foundational questions posed by Alan Turing to the modern era of AI research. We categorize the evolution of LLMs into distinct periods, each characterized by its unique benchmarks and evaluation criteria. As LLMs increasingly mimic human-like behaviors, traditional evaluation proxies, such as the Turing test, have become less reliable. We emphasize the pressing need for a unified evaluation system, given the broader societal implications of these models. Through an analysis of common evaluation methodologies, we advocate for a qualitative shift in assessment approaches, underscoring the importance of standardization and objective criteria. This work serves as a call for the AI community to collaboratively address the challenges of LLM evaluation, ensuring their reliability, fairness, and societal benefit.
\end{abstract}

\section{Introduction}

Alan Turing began his famous article "Computing Machinery and Intelligence" \cite{turing1950computing} by stating that it is extremely difficult to formulate objective definitions of the terms "machine" and "think" in the context of the question: \textit{Can machines think?}
Instead, he proposed looking for an answer to another question: \textit{Can machines reliably imitate human dialogue?}

Back then, in 1950, the answers to both questions were so far apart from us that the difference between them was insignificant, and this substitution helped to set the "north star metric" for a long time, the direction of development for the entire field of research, including dialog systems, human-machine interfaces, and various kinds of AI. A possible reason for this success is that a practical solution to this imitation task implies the need to fulfill (to some extent) several complex conditions simultaneously, including natural language proficiency, interactivity, and effective grasp on the context of the conversation. Moreover, since the initial setup does not specify the fixed protocol, other strong requirements may be implied, such as common knowledge of the world, reasoning, abstract or creative thinking, concept of causality, and so on, depending on the particular interviewer's questions.

Now, 73 years after Turing's paper, modern systems have greatly evolved, successfully mimicking human-like behaviors and interactions. The first officially documented machine passed the Turing test in 2014 \cite{Warwick}, long before the era of Large Language Models. Since then, the quality of dialog simulation and text generation in general has increased even more, so the Turing test has long since ceased to serve as a reliable proxy for evaluation of modern systems. Instead, a wide variety of approaches are used in practice, aimed to assess different individual abilities and properties of a system. However, we have neither a unified system of criteria nor clear formulation of the evaluation goals. In the meantime this new evaluation methodology will not only influence the trajectory of AI research but will also have broader implications. Thus, it is paramount to ensure that LLMs are reliable, unbiased, and beneficial for society.

This paper does not set a general goal for further development of LLMs but tries to provide a comprehensive overview of the evaluation methodologies for Large Language Models and dialog agents. In Section 3, we present a chronological overview of the recent history of LLM development and their evaluation methods. Specifically, we explore benchmarks, human assessments, and model assessments, among others, that are prevalent in both academic research and practical applications. In Section 4, we propose a primary taxonomy of these approaches and discuss their strengths and internal issues, including noticeable errors, problems, and contradictions. Section 5 examines which specific aspects of LLMs are commonly evaluated in contemporary studies. Finally, in Section 6 we use the proposed taxonomy to discuss current challenges and possible directions for further progress in the field.

One has to state that the current evaluation approaches have are not effective and do not meet modern requirements. Moreover, further extensive development of the existing approaches (for example, increasing the number of benchmarks and creating new tasks within existing benchmarks) cannot address these issues. We drastically need a qualitative rather than quantitative leap in evaluation. In our opinion, the first step towards a solution should be the survey of the existing evaluation taxonomy, and a detailed discussions of the weaknesses of the available methods that we try to provide in this paper.



\section{"Cambrian explosion" of large language models}
\begin{figure}[t]
    \centering
    \includegraphics[scale=0.35]{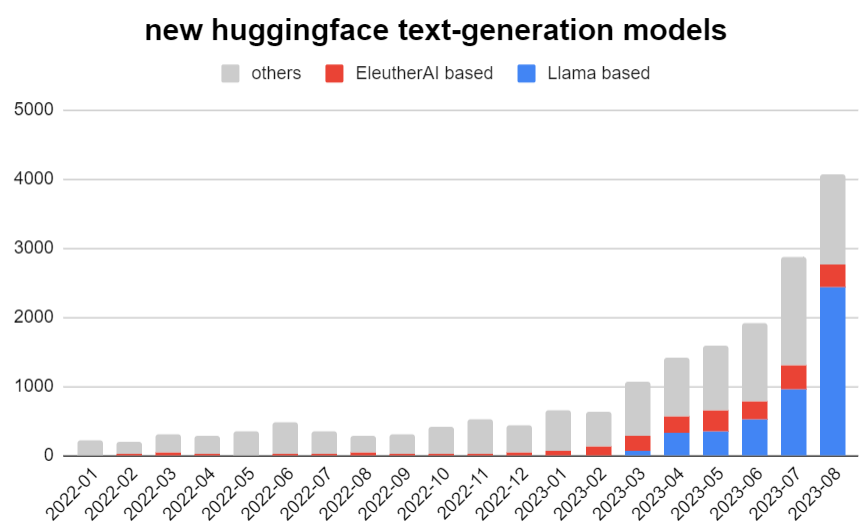}
    \caption[Cambrian explosion of large language models: the number of monthly created text-generation model repositories on huggingface, based on statistics by HFCommunity.]{Cambrian explosion of large language models: the number of monthly created text-generation model repositories on huggingface, based on statistics by HFCommunity.\protect\footnotemark}
    \label{fig:explosion}
\end{figure}

\footnotetext{\url{https://som-research.github.io/HFCommunity/index.html}}

Lately the landscape of language models has expanded remarkably (Figure \ref{fig:explosion}). As of October 2023, the number of generative text models on Hugging Face (HF) has reached a remarkable 25 000+ and 86 59
models are based explicitly on the LLaMA model \cite{touvron2023llama}. This explosion can also be observed in real time\footnote{For example, visit \url{https://github.com/hollobit/GenAI\_LLM\_timeline}}.

With such an abundance of models, it becomes essential to evaluate and compare their quality. A state-of-the-art survey by \citealp{yang2023harnessing} provides valuable insights into the diverse applications and capabilities of language models beyond ChatGPT. However, the various works in this field employ different methodologies for assessing quality. Expansion at such a rate brings inevitable confusion\footnote{For example, the very term "large language models" is constantly used, but there is no universally accepted threshold for the number of parameters after which the model is considered large.} within the field. So, common evaluation methodologies are not only far from consistent but are also contradictory sometime.

\begin{figure}[t]
    \centering
    \includegraphics[scale=0.35]{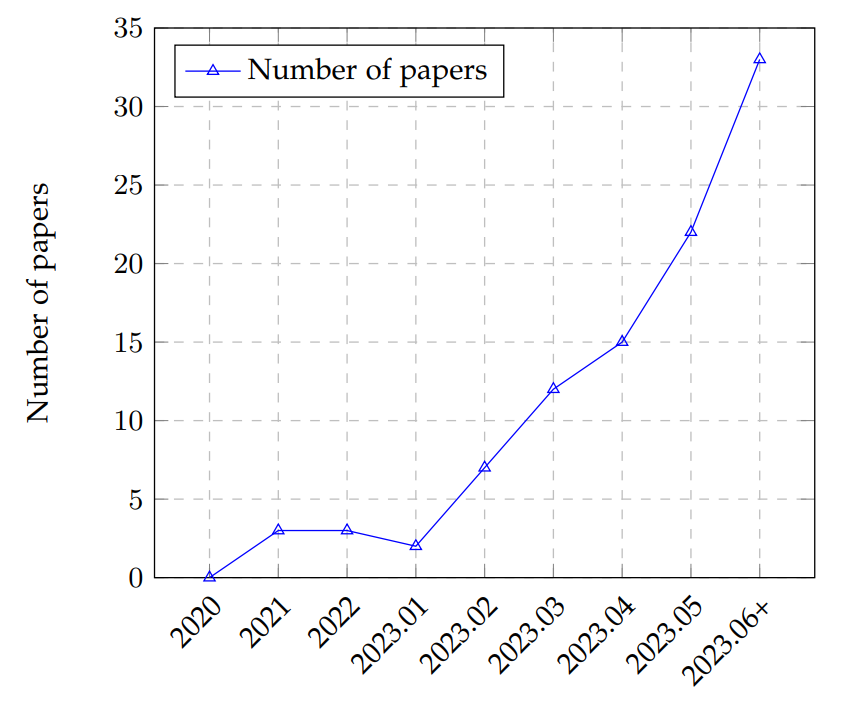}
    \caption{Trend of LLMs evaluation papers over time from \citealp{chang2023survey}}
    \label{fig:papers}
\end{figure}

This paper has no intention to provide a complete and comprehensive survey of the field. We suggest focusing on one aspect of LLM development that we personally see as the most crucial for the future progress of the field, namely, evaluation. However, even in this narrowed context, it is hardly possible to guarantee any form of a complete review due to the number of relevant papers on the topic (Figure \ref{fig:papers}). We address the reader to \citealp{chang2023survey} for an example of such a survey. In this paper, we instead discuss selected examples to illustrate the trends and challenges we are facing. We believe those examples are relevant to the field and had a high impact at the moment of their release. We do not claim we can provide a full review of all evaluation techniques used for LLMs, but to the best of our knowledge, this paper lists all significant conceptual approaches.

\section{Evolution of LLM Evaluation}

Let us review the trends in LLM Evaluation. Subjectively, we split LLM development into three core periods with specific properties. We list some of the models for every period and briefly describe the methods used for performance evaluation. We do not imply that the list of the models is complete. We also list only some of the evaluation methods used for every model since they are numerous and tend to overlap. Nevertheless, we enumerate the primary evaluation methodologies so the reader can have a fair and complete representation of the spectrum of evaluation methods available today. Let us briefly discuss each period and highlight some of the methods that were used for evaluation.

\subsection{"Prehistoric" LLM Evaluations}

In this subsection, we discuss evaluations of models that emerged before the appearance of GPT-3\footnote{\url{https://openai.com/blog/gpt-3-apps/}}, which was initially released in beta on June 11, 2020. We have mentioned above that there is no consensus on the threshold for the "large" language model. Thus, we suggest discussing models with more than one billion parameters\footnote{Appendix contains Table \ref{tab:1} with comprehensive overview of all core models discussed in the paper}. 








During this period, the models are mainly assessed on relatively simple and common NLU benchmarks such as LAMBADA \cite{paperno2016lambada}, GLUE \cite{wang2018glue}, SuperGLUE \cite{wang2019superglue}, SQuAD \cite{rajpurkar2016squad}, MNLI \cite{williams2018broad}, QQP \cite{wang2017bilateral}, SQuAD, Winograd Schema Challenge \cite{levesque2012winograd}, RACE \cite{lai2017large}, or similiar. Since LLMs from this period achieved at most 50\%-80\% of human-level performance on these tasks, the progress across various models was clearly visible.  In some papers, the authors try to devise additional metrics for model performance comparison. For example, several papers compare the perplexities using the same WikiText dataset, which is questionable since models often have different tokenization vocabularies. Hence, comparing such perplexities could only be fair with some additional tricks (see, for example, \citealp{mosin2023fine}).

\subsection{From GPT-3 to ChatGPT}

During this period, before the end of 2022, the number of new LLMs has increased\footnote{See Table \ref{tab:1}B}, since several major developers joined the race. These new models consistently achieved scores of 90\% or higher on some of the old benchmarks (e.g., SuperGLUE, LAMBADA, SQuAD, GLUE), so they became less informative because of limitations of their sensitivity. 

Consequently, researchers tend to use more complex and/or specific benchmarks, such as StoryCloze \cite{mostafazadeh2017lsdsem}, HellaSwag \cite{zellers2019hellaswag}, TriviaQA \cite{joshi2017triviaqa}, ARC \cite{clark2018think}, CoQA \cite{reddy2019coqa}, DROP \cite{dua2019drop}, QuAC \cite{choi2018quac}, SQuADv2 \cite{Rajpurkar2018SQuAD2}, hoping to capture nuances of different models' quality. 

Moreover, new complex benchmarks (such as PIQA \cite{bisk2020piqa} and  
Closed Book Question Answering \cite{wang2021can}) were introduced. Notably, benchmarks such as MMLU \cite{hendrycksmeasuring}, BIG-Bench \cite{srivastava2022beyond} as well as HELM meta benchmark \cite{liang2022holistic}, often covering multiple disciplines akin to a human exam, have emerged as evaluation tools. 

However, there is no universally agreed-upon system of benchmarks, leading to arbitrary comparisons across various evaluation criteria. At the same time, such an abundance of comparison scales leads to the absence of Pareto superiority for any given model\footnote{Pareto superiority is as a situation when a new model outperforms the previous ones on all evaluation tasks.}. Instead, authors now commonly state, "\textit{our model outperforms the prior state-of-the-art on X out of Y tasks}."

Another essential trend of this period is the wide usage of human labeling primarily used to deal with specific or subjective aspects of evaluation. Since the costs of high-quality human labeling are high, using an analog of the chess Elo rating, known as ELO \cite{arpad1978rating}, established itself as a potential solution for sparse pairwise comparisons.

During this period, researchers attempt to assess the toxicity, biases, and harmful behavior of LLMs, using dedicated benchmarks together with human evaluation. In this paper, we deliberately do not discuss toxicity assessment or alignment issues, as this is a separate significant topic for which we refer to \citealp{sorensen2023value}.

\subsection{Modern Era}


Finally, we would like to highlight notable language models released in 2023 (Table \ref{tab:1}C) and provide details about their evaluations. 

The introduction of open models such as LLaMA and Pythia \cite{biderman2023pythia}, among others, has significantly increased the number of researchers conducting experiments with LLMs. Since the number of models is rising exponentially, see Figure \ref{tab:1}, probably, a couple of new models appeared just while you read this paper. We have no intent to enumerate all available LLMs; instead, we try to capture the main trends and patterns here:

\begin{itemize}
    \item the development and heavy usage of various complex benchmarks continues,
    \item many new evaluations are based on human school exams or other tests initially designed for humans, such as GMAT, SAT, LSAT, etc.
    \item toxicity/bias/hate speech assessments (as well as helpfulness, honesty, and harmlessness)  become a mandatory attribute of the overall model evaluation,
    \item the complexity of the evaluation criteria motivates researchers to use pairwise evaluation when possible,
    \item high costs of pairwise labeling lead to the extensive use of other, superior models (mainly ChatGPT or GPT-4) for evaluation,
    \item these sparse pairwise or side-by-side evaluations, combined with an Elo rating system, enable the creation of leaderboards for model comparison.
\end{itemize}

Another trend worth mentioning is the rise of code-generation LLMs since they have significant specifics in application and evaluation approaches. We mention just some of them, including StarCoder \cite{li2023starcoder}, CodeGeeX \cite{zheng2023codegeex}, and WizardCoder \cite{luo2023wizardcoder}. Such models usually utilize special benchmarks with auto-tests for generated code (including HumanEval \cite{chen2021evaluating}, HumanEval+ \cite{liu2023your}, DS-1000 \cite{lai2022ds}, or MBPP \cite{austin2021program}).

\section{Prevalent Evaluation Methodologies}

As the field evolved, several generalized approaches to evaluation established themselves. These include comparing the models on a set of benchmarks, assessment by humans, and modeling human evaluation (either using heuristics, dedicated models, or a superior LLM model). Each of these approaches has its advantages, limitations, and potential drawbacks. Let us analyze them sequentially to understand their specifics.

\subsection{Comparison on benchmarks}

Benchmarks may provide a fast and reliable evaluation of models. In some sense, benchmark evaluation resembles commonly used tests for human performance evaluation. The critical requirements here are the standardization of test sets and the controlled environment of evaluation. There are several interesting developments towards standardization such as HELM\footnote{\url{https://crfm.stanford.edu/helm}}, BIG-Bench\footnote{\url{https://github.com/google/BIG-bench}} or \citealp{eval-harness}. The last one makes an interesting step to provide a unified benchmarking framework that includes 200+ tasks for evaluation and supports a variety of available LLMs. 

At the same time, similarly to human tests, LLM benchmarks have disadvantages:
\begin{itemize}
    \item While we are in the active phase of LLM quality improvement, old benchmarks become obsolete quickly; however, they are often still included in the evaluation procedures.
    \item Since new benchmarks are not fully standardized yet, they often overlap or contradict, which may lead to some inconsistency. 
    \item Taking into account the low number of tasks per topic (for example, MMLU consists of 57 types of questions on mathematics, history, psychology, etc., with an average of 280 questions per topic), the randomness may affect the outcome for each topic a lot. For example, it was shown that minor changes in the multiple-choice formatting can cause a performance jump of 6-10 points on MMLU\footnote{\url{https://twitter.com/ArmenAgha/status/1669084129261162497}}. The standard way to deal with noise is to measure confidence intervals; however, the limited data available does not enable the use of bucket test statistics.
    \item A tempting idea for noise control is averaging results across several different independent benchmarks and publishing the resulting ratings\footnote{See some examples: \url{https://huggingface.co/spaces/HuggingFaceH4/open_llm_leaderboard},
 \url{https://shorturl.at/DGPW3}, \url{https://github.com/FranxYao/chain-of-thought-hub}, \url{https://cevalbenchmark.com/static/leaderboard.html}, \url{https://bellard.org/ts_server/}, \url{https://huggingface.co/spaces/toloka/open-llm-leaderboard}
 }. However, the resulting rating often fails to account for possible methodological flaws or deliver a tangible value to a larger NLP community \cite{ethayarajh2020utility}. 
    \item The known problem of standardized benchmark evaluations is leakage or so-called \textit{test set pollution} since some of the benchmarks have been available on the internet for years (e.g., MMLU since 2021) and can easily occur in pre-training or fine-tuning datasets. A couple of such recent high-profile cases have sparked heated discussion in the community\footnote{Check, for instance, \url{https://huggingface.co/spaces/HuggingFaceH4/open_llm_leaderboard/discussions/213}}, and led to criticism in satirical papers like \citealp{schaeffer2023pretraining}.
    \item Another known issue of modern benchmarking is its massive computational costs: benchmarks typically have the order of $10^5$ validation examples, with $10^3$ - $10^4$ per task, extending the load up to hundreds of GPU hours per model evaluation. Some recent works, like \citealp{vivek2023anchor} and \citealp{perlitz2023efficient}, try to reduce these computational costs, but it is still hard to keep the reasonable stability of results simultaneously.
    \item Also, as we mentioned before, reducing the number of test topics or tasks may be dangerous in terms of intended or unintended cherry-picking, making it easy to choose the ones where a particular model performs well.
\end{itemize}

Summing up, using benchmarks is a good starting point for rough evaluation. However,  benchmarks have several significant drawbacks, including insufficient standardization, high computational costs, poor robustness to noise, and frequent cases of test set leakage. Moreover, benchmark assessments often do not agree with the human assessment of the model performance\footnote{Some examples of such inconsistency are available at \url{https://llm-leaderboard.streamlit.app/} or \url{https://github.com/LudwigStumpp/llm-leaderboard}}, making, potentially, the whole evaluation inconsistent. Let us now discuss the human evaluation more thoroughly.

\subsection{Evaluation by Human Assessors}

Evaluation by human assessors is an expensive yet widely used approach. 
While it may be possible to train and use a dedicated model for almost any well-formulated aspect of evaluation, the core problem is precisely in formulating a detailed definition of the evaluation criteria. The typical way to evade this is by asking about assessors' overall preference in a pairwise (side-by-side) setup and then building a rating between available models or configurations based on these pairwise scores. However, this workaround comes with its own set of challenges and drawbacks.

First, the complete pairwise evaluation is too expensive and time-consuming to compare a significant number of models since the complexity of the procedure grows like O($n^2$) with the number of compared models.

Second, pairwise comparisons can yield non-transitive results, making it challenging to establish a consistent global ranking. In other words, without clearly articulated criteria, human assessors may prefer system A to system B, system B to system C, and system C to system A. Researchers use different methods to deal with such situations. One alternative could be Elo rating\footnote{Elo ratings have their own limitations discussed in \cite{szczecinski2020understanding}.} or relative comparison of evaluated models with one clearly weaker LLM. For an example of a more advanced ranking method, see \citealp{NEURIPS2022_020e313d}.

On the other hand, numerous co-existing leaderboards\footnote{Examples include \url{https://chat.lmsys.org/?leaderboard}, \url{https://github.com/LudwigStumpp/llm-leaderboard}, \url{https://aviary.anyscale.com/}, and \url{https://llm-leaderboard.streamlit.app/}} may provide different rankings for the same models since they are based on different sets of noisy human pairwise labels, while the noise measurements and confidence intervals are usually absent due to the low amount of data.


Another significant issue is the quality of human labels, which can be relatively low for different reasons. Human assessors' motivation is sometimes insufficient to provide high-quality answers; moreover, some assessors secretly use LLMs as to speed up the labelling \cite{veselovsky2023artificial}. This might introduce unexpected shifts in the obtained assessments. Furthermore, the absence of global criteria may lead to situations when human assessors prefer more good-looking and stylish responses to correct and factual ones \cite{gudibande2023false}.



Since the research community tend to treat human assessment as an expensive ground truth, researchers often try to model human evaluation with heuristics or some dedicated algorithm to reduce the evaluation's complexity and cost. Let us discuss these methods in the following subsection.

\subsection{Modeling Human Evaluation}

One of the common ways to obtain a cheaper estimation of human assessment is to train a dedicated model on existing human labels to predict them and then use it as a replacement for human assessors. Dozens of such approaches are proposed; for example, in the domain of dialog agents evaluation there are methods like FED \cite{mehri2020unsupervised}, USL \cite{phy2020deconstruct}, Flowscore \cite{li2021conversations}, QuestEval \cite{scialom2021questeval}, Open AI detector\footnote{\url{https://huggingface.co/roberta-base-openai-detector}}, CT Score\footnote{\url{https://github.com/tanyuqian/ctc-gen-eval}}, FULL score \cite{de2022open}, Reranker\footnote{\url{https://github.com/luyug/Reranker}}, Cross-Encoder\footnote{\url{https://huggingface.co/cross-encoder/ms-marco-MiniLM-L-6-v2}} for MS-Macro\footnote{\url{https://github.com/microsoft/MSMARCO-Passage-Ranking}}, Quality Adapt \cite{mendonca-etal-2022-qualityadapt}, Deam score \cite{ghazarian2022deam}, RankGen \cite{krishna2022rankgen} and many others.

Although successfully implementing a human preferences model is usually necessary for the RLHF to have the so-called \textit{Reward Modeling}, there is still no ultimate solution. However, the situation has changed significantly with the appearance of modern LLMs since one can compare the outputs of to models using a superior one.

As of today, GPT-4 is the most prominent candidate for such a superior model, which can be used (see, for example, \citealp{zheng2023judging}) to evaluate or compare the candidates instead of humans without additional fine-tuning. Moreover, \citealp{thomas2023large} reports that GPT-4 produces better relevance labels than third-party workers. However, even GPT-4 has a couple of known significant issues, including:

\begin{itemize}
    \item GPT-4 is also known to have a specific vocabulary bias, particularly it prefers its own generations more than humans do \cite{zhou2023lima},
    \item GPT-4 seems to have specific positional biases\footnote{\url{https://twitter.com/nazneenrajani/status/1667224735573487616}},
    \item Some systematic contradictions between GPT-4 and human assessment are reported \cite{xu2023wizardlm},
    \item GPT-4 biases may be misaligned with human biases, which makes the idea of the blind comparison by a GPT-4 model quite challenging. 
\end{itemize}

Such problems are not specific to GPT-4 but appear in the results of different models in different ways. The recent paper on the CoBBLEr benchmark \cite{koo2023benchmarking} studies these effects across 15 existing LLMs.

Overall, it seems like we cannot avoid a clear definition of what we are evaluating without introducing significant noise or bias into the results.

\section{What Are We Evaluating?}

\begin{figure*}[t]
    \centering
    \includegraphics[scale=0.42]{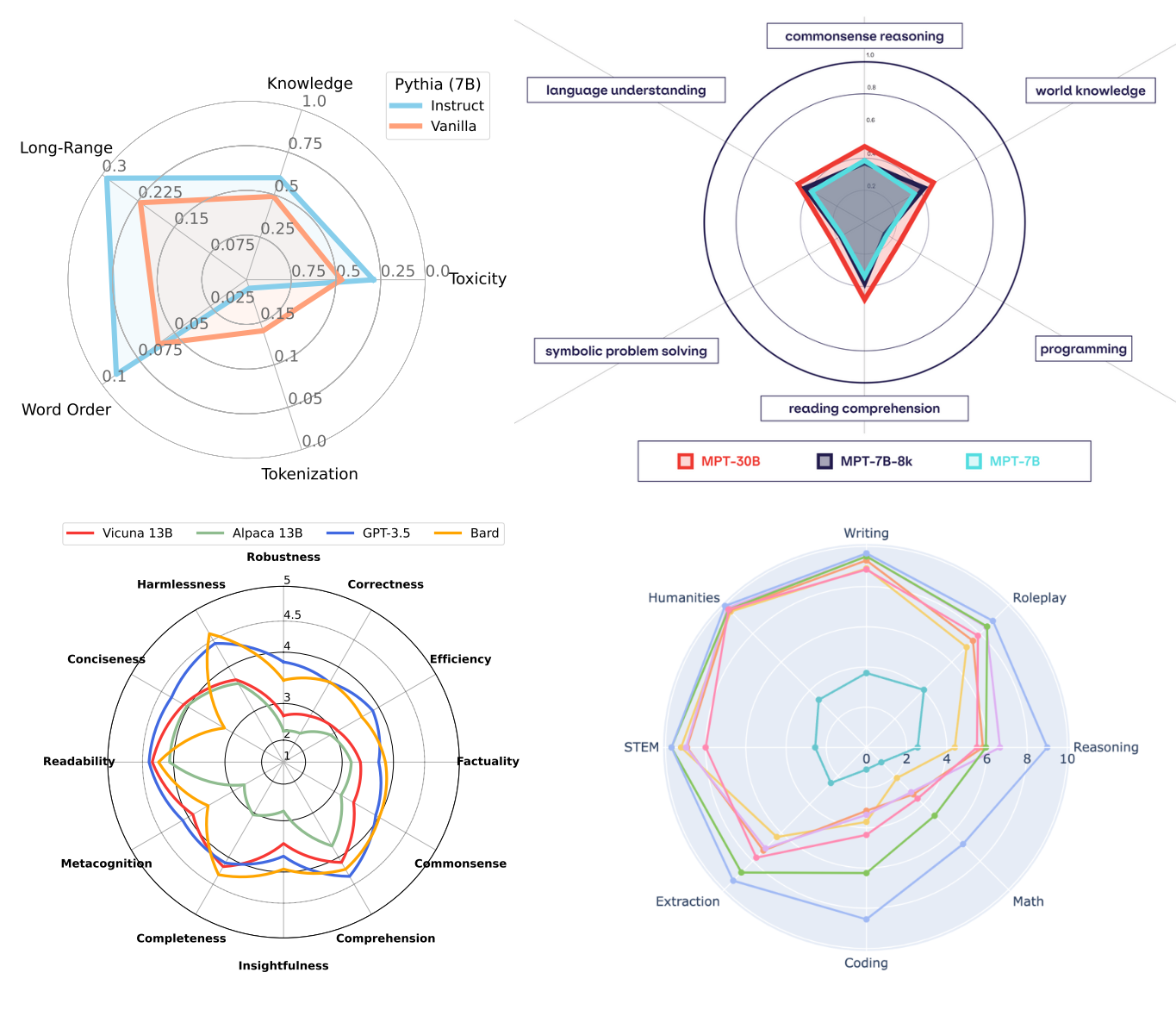}
    \caption{Radar diagrams for several recent models. Top-left is from \cite{jain2023bring}, top-right is from  Mosaic Eval Gauntlet, 
    bottom-left is from \cite{ye2023flask}, bottom-right is from the Giraffe-70b release.
    }
    \label{fig:radars}
\end{figure*}


With dozens of actively used benchmarks with hundreds of task types, researchers naturally tend to group them into general aspects of the model's performance, so providing several high-level scores becomes standard practice. Often, researchers present them as so-called \textit{radar diagrams} to highlight the advantages and disadvantages of the given model over baselines.

However, an overview of recent papers reveals no structure or system of these aspects, even on the highest level (see Figure \ref{fig:radars}). Sometimes, they remind the famous fiction animals classification \cite{Borges}, mixing different types and principles altogether. Building a proper taxonomy for these aspects is a complex and extensive endeavor, far beyond the scope of this paper. For deeper insights on this topic, we address the reader, for example, to \citealp{ziyu2023through} or \citealp{xuanfan2023systematic}. Here, we just mention some commonly used approaches and group them intuitively, then discuss the results.

\begin{itemize}
\item \textbf{Text-specific and dialog-specific abilities} are crucial since textual dialogues are the common medium for modern LLMs. They may include:
\begin{itemize}
\item General text comprehension and natural language understanding (for example, LAMBADA benchmark);
\item Multilingualism (many options, including recently published BELEBELE \cite{bandarkar2023belebele});
\item Plausibility of dialog communication;
\item Capability to understand and control the text quality, style, and level of details;
\end{itemize}
\item \textbf{Knowledge-specific characteristics} - characteristics of knowledge obtained by the model during training:
\begin{itemize}
\item Common knowledge is essential since human communication is built on the existence of implicitly shared contexts \cite{ClarkB91};
\item Depending on the context or application, we may want to assess models' niche knowledge, such as Humanities or STEM; benchmarks here are usually compiled based on human exams or manually crafted tests like BIG-Bench;
\end{itemize}
\item \textbf{Skill-specific abilities} - abilities to solve problems that require some skills besides knowledge:
\begin{itemize}
\item Commonsense reasoning\footnote{See a survey on Commonsense Reasoning benchmarks in (\citealp{davis2023benchmarks})};
\item Abstract reasoning and ability to generalize\footnote{(\citealp{chollet2019measure}) proposes to assess reasoning without modulation by prior knowledge and experience};
\item Specific skills (Code generation, Roleplay, Math reasoning, Image manipulation, Chess problem solving, etc.)\footnote{There are many specific skills benchmarks, see, for example, the recent NuclearQA bencmark \cite{acharya2023nuclearqa} or the RoleLLM framework \cite{wang2023rolellm}};
\end{itemize}
\item \textbf{Personality and CogSci features} - since the general modern models' UI is a dialog via chat, users and researchers tend to treat them as personalities; this leads to the idea of corresponding attributes measurement:
\begin{itemize}
\item Creativity\footnote{\url{https://bit.ly/3rKZWLm}}, Empathy, Emotional Intelligence \cite{wang2023emotional}, or Social awareness \cite{zhan2023socialdial}; 
\item Cognitive Science-related aspects include planning and cognitive mapping abilities \cite{momennejad2023evaluating}, deductive competence \cite{seals2023evaluating}, and complex reasoning skills \cite{kuo2023large};
\end{itemize}
\item \textbf{Alignment, Reliability, and Safety related features}, including
\begin{itemize}
    \item Alignment to human values\footnote{See the survey by \citealp{yao2023instructions}};
    \item Security, which encompasses various aspects, like privacy, preventing malicious use, and addressing potential biases;
    \item H4 attributes\footnote{\url{https://huggingface.co/HuggingFaceH4}}, namely being Helpful, Honest, Harmless, and Huggy, reflecting positive social qualities;
    \item Factuality \cite{chen2023felm}, truthfulness, and the ability to acknowledge uncertainty or lack of knowledge;
    \item Explainability\footnote{Though, \citealp{hsia2023goodhart} recently showed the flaws of available explainability metrics.};
\end{itemize}
\item \textbf{Technical characteristics} (including Long-range context \cite{dong2023bamboo}, tokenization quality, etc)

\end{itemize}

These diverse evaluation dimensions highlight the multifaceted nature of assessing language models, each with unique considerations and challenges. For example, the precise definition of text style remains challenging \cite{tikhonov2018wrong}, while storytelling evaluation needs a deeper understanding of the concept of narrative \cite{gervas2019long,yamshchikov-tikhonov-2023-wrong}. Indeed, the evaluation guidelines proposed in \cite{hamalainen2021human} for creative, generative systems are relevant for the LLM evaluation in general: \textit{"clearly defining the goal of the generative system, asking questions as concrete as possible, testing the evaluation setup, using multiple different evaluation setups, reporting the entire evaluation process and potential biases clearly, and finally analyzing the evaluation results more profoundly than merely reporting the most typical statistics."}

 A well-defined and structured list of aspects we want to evaluate LLM on is essential to optimize and prioritize the evaluation of language models. Do we really need them all? How do they interrelate? Without a clear understanding of what aspects we are assessing and why, it becomes difficult to focus on specific areas for improvement or to allocate resources effectively. 

\section{Discussion}

Let us now try to sketch the main trends in evaluation approaches and hypothesize their further development in the context of the multiple challenges we highlighted above. 

 \subsection{Human-like Evaluation}

It is worth noting that most of the current approaches to model evaluation listed in this paper are essentially anthropocentric.
One reason for this may be that benchmarks are opportunity-driven. Instead of creating new, specifically targeted tests, many researchers adapt existing ones created for humans in the past. 

At first glance, this simplifies not only their creation but also the interpretation of results. However, some of these tests are designed specifically for assessing human adults and might not be well suited for evaluating a broader range of signatures of intelligent behavior \cite{Eisenstein2023}.

Another disadvantage of this approach is that it may limit the assessment scale. Now, when superhuman performance has been achieved in some tasks, this may become a constraint or an extra incentive that distorts goal setting. For example, the need to pass a classical Turing test may encourage a model to deceive the tester and hide part of its abilities (as it may be given away by too high a calculation speed or too deep an encyclopedic knowledge).

Suppose we want to drive and track the development of models' abilities at levels qualitatively higher than the current humans. In that case, we should consider creating fundamentally new approaches, for example, developing particular competitive evaluation environments that assess not built-in knowledge and abilities but the speed and quality of forming new skills in an interactive, unfamiliar environment. We see the ARC benchmark from \citealp{chollet2019measure} as a good step in this direction.

 \subsection{Decompose and Conquer}

However, there is one thing we might want to use from the experience of human skills testing. Just like human IQ test are split into several subcategories, like Short-Term Memory, Reasoning, and Verbal \cite{Hampshire2012}, we need to divide potential LLM skills into a standardized system and define generic baselines.

There still are debates about whether it is possible to develop a universal measure of intelligence.  In the meantime, we clearly see the progress of LLMs across specifically defined tasks. With limited resources and various practical tasks, developers may not want to build universally superior models. Instead, they can focus on the selected skills and abilities. For example, creators of a code assistant should not bother themselves with improving the literature style of their model too much. We believe that this tactic of "decompose and conquer" will further dominate the field, so making the rules, requirements, and systematic baselines global and public should benefit the whole community.





 \subsection{Nobody's Perfect}

Another interesting observation is that we tend to perceive and evaluate modern models as agents in communication with humans. We earnestly expect LLMs to behave in a socially acceptable way -- imposing requirements like factuality, harmlessness, helpfulness, etc.

For some parameters, we impose stricter requirements on the evaluated models than we would if we were evaluating ordinary people (e.g., we may allow some sloppiness, inattention, or carelessness from a living person, but we require models to be free of such problems). These strong demands might be rooted in the fact that we already use such models to create mass services in which they act as experts in some narrow field (data processing, science, medicine, law, etc).

Accordingly, we already expect LLMs to have confident and stable expert knowledge and skills in the target domain, implying that requirements like natural language skills and the ability to maintain a conversation are self-evident. This perfectionist bias appears likely to stay with us and potentially intensify, as testing specific skills in models will become increasingly complex and expensive.

 \subsection{Independent Evaluation Bodies}

 The evaluation and certification of LLMs could be a separate field in itself. Indeed, various global organizations work on evaluations of various human cognitive skills. There is no reason why a similar pattern could not emerge for LLMs. Creating efficient leak-proof test methodologies will only be more demanding as the models progress. At the same time, for-profit organizations clearly need some form of evaluation to compare their solutions with the competition. This might create a market incentive for the creation of for-profit organizations that could be centered around LLM certification and evaluation. 

 \section{Conclusion}

This paper provides an overview of the current state of evaluation techniques used for LLMs and analyzes them. We trace the progress of LLMs in the last few years and create a taxonomy of the methods used to evaluate LLM performance. One by one, we analyze significant approaches and highlight challenges that arise with them, including insufficient standardization, poor robustness to noise, and test set leakage of benchmarks; frequent cases of disagreement between benchmark-based evaluations, humans' and superior models' preferences; humans' and superior models' biases; dead ends of Pareto optimization and non-transitive results in the absence of global criteria; no structure or system of aspects of evaluations, even on the highest level.

Based on these observations, the current evaluation approaches have lost their effectiveness and do not meet modern requirements, and there is no clear way to patch them. In our opinion, the first step towards a solution should be the standardization of tasks and evaluation methods, including a precise formulation of the assessed aspects. We still do not know whether there is a new single "Turing question" that can set the main direction of the industry for the following decades. What is certain is that to figure out how to move forward, we need to precisely articulate what we want to measure and for what reason.

\bibliography{anthology,custom}
\bibliographystyle{acl_natbib}

\appendix
 \section*{Appendix}
 
\begin{table*}[hp]
  \centering
  \begin{threeparttable}  
  \begin{tabular}{|l|m{25em}|}
  \hline
  \multicolumn{2}{|c|}{\small \textbf{A. The "prehistoric" era of LLM}} \\
  \hline
  \small 2019, GPT-2\tnote{a} 
  & \small LAMBADA, WSC, QA
, summarization, translation tasks, etc. \\ 
  \hline
  \small 2019, T5 \cite{raffel2020exploring} & \small GLUE, SuperGLUE,  SQuAD, QA, summarization, translation tasks, etc. \\
  \hline
  \small 2019, CTRL \cite{keskar2019ctrl} & \small no include explicit quality measurements. \\
  \hline
  \small 2019, Megatron-LM \cite{shoeybi2019megatron} & \small LAMBADA, MNLI, QQP, SQuAD, RACE, etc. \\
  \hline
  \small 2020, Turing-NLG\tnote{b} & \small LAMBADA, summarization, etc. \\
  \hline
  \multicolumn{2}{|c|}{\small \textbf{B. From GPT-3 to ChatGPT}} \\
  \hline
  \small 2020, GPT-3\cite{gpt3} & \small LAMBADA, StoryCloze, HellaSwag, Closed Book Question Answering, TriviaQA, PIQA, ARC, CoQA, DROP, QuAC, SQuADv2, RACE, SuperGLUE, NLI, OpenBookQA, some other tasks inspired by human school exams, and human side-by-side evaluation. \\ 
  \hline
  \small 2021, Blenderbot \cite{shuster2021multi} & \small human side-by-side evaluation. \\
  \hline
  \small 2021, Gopher \cite{rae2021scaling} & \small 152 diverse tasks from different benchmarks, including LAMBADA, MMLU, BIG-bench, TriviaQA, NaturalQuestions, TruthfulQA, PIQA, WinoGrande, SocialIQA, HellaSwag, plus some tasks inspired by human school exams, plus some toxicity, bias and hate speech evaluation. \\
  \hline
  \small 2021, GLaM \cite{du2022glam} & \small compared to GPT-3 and Gopher across 29 benchmarks. \\
  \hline
  \small 2022, OPT \cite{zhang2022opt} & \small compared to GPT-3 across 16 tasks, plus some toxicity, bias and hate speech evaluation. \\
  \hline
  \small 2022, LaMDA \cite{thoppilan2022lamda} & \small human assessments on specific aspects, including sensibleness, specificity, interestingness, safety, and factual grounding. \\
  \hline
  \small 2022, PaLM \cite{chowdhery2022palm} & \small evaluated on 29 benchmarks, which were similar to the set of tasks used for GPT-3 + MMLU and BIG-Bench. \\
  \hline
  \small 2022, Chinchilla \cite{hoffmann2022training,hoffmann2022empirical} & \small benchmarks included MMLU, BIG-bench, and other. \\
  \hline
  \small 2022, BLOOM \cite{scao2022bloom} & \small 20 benchmarks, which were a subset of those used for GPT-3. \\
  \hline
  \small 2022, InstructGPT\tnote{c} & \small human assessments of specific aspects, used Elo rating. \\
  \hline
  \small 2022, ChatGPT\tnote{d} & \small evaluations were conducted based on InstructGPT. \\
  \hline

  \multicolumn{2}{|c|}{\small \textbf{C. The "modern" era}} \\
  
  \hline
  \small 2023, GPT-4\tnote{e} & \small benchmarks including MMLU, HellaSwag, WinoGrande, and others + academic and professional examinations. \\ 
  \hline
  \small 2023, LLaMA \cite{touvron2023llama} & \small MMLU, HellaSwag, WinoGrande, ARC, and more. \\
  \hline
  \small 2023, Alpaca \cite{taori2023alpaca} & \small minimal evaluation. \\
  \hline
  \small 2023, Claude\tnote{f} & \small minimal evaluation. \\
  \hline
  \small 2023, Vicuna \cite{vicuna2023} &  \small side-by-side compared to Alpaca and LLaMa by GPT-4  as a judge. \\
  \hline
  \small 2023, WizardLM \cite{xu2023wizardlm} & \small side-by-side assessment by human evaluators and GPT-4. \\
  \hline
  \small 2023, MPT family of models\tnote{g} & \small several standard benchmarks + code specific tasks, like HumanEval. \\
  \hline
  \small 2023, Palm-2 \cite{anil2023palm} & \small similar to GPT-4 - a lot of standard benchmarks (including, for example, BIG-Bench and Winogrande) + language proficiency exams. \\
  \hline
  \small 2023, Claude-2\tnote{h} & \small benchmarks, alignment, lanugages, long context. \\
  \hline
  \small 2023, Falcon \cite{falcon40b} & \small standard benchmarks, including ARC, HellaSwag, MMLU, TruthfulQA. \\  
  \hline

  \end{tabular}
    \begin{tablenotes}
    \small \item [a] \url{https://openai.com/research/better-language-models}
    \small \item [b] \url{https://shorturl.at/epK79}
    \small \item [c] \url{https://openai.com/research/instruction-following}
    \small \item [d] \url{https://openai.com/blog/chatgpt/}
    \small \item [e] \url{https://openai.com/gpt-4}
    \small \item [f] \url{https://www.anthropic.com/index/introducing-claude}
    \small \item [g] \url{https://github.com/mosaicml/llm-foundry}
    \small \item [h] \url{https://www.anthropic.com/index/claude-2}
    
    \end{tablenotes}  
  \end{threeparttable}  
  \caption{Selected examples of LLM Evaluation approaches}
  \label{tab:1}
\end{table*}

\end{document}